\documentclass[10pt, a4paper]{article}

\usepackage{lrec-coling2024} 

\usepackage{amsmath}
\usepackage{amssymb}
\usepackage{xstring}
\usepackage{bbding}  
\usepackage{multirow}
\usepackage{color}

\usepackage{booktabs}

\usepackage{fontawesome}
\newcommand{\tox}{\faExclamationTriangle}
\newcommand{\ton}{\faHandPeaceO}

\title{Take its Essence, Discard its Dross! Debiasing for \\ Toxic Language Detection via Counterfactual Causal Effect }

\name{Junyu Lu$^{\dag}$, Bo Xu$^{\dag}$, Xiaokun Zhang$^{\dag}$, Kaiyuan Liu$^{\dag}$, Dongyu Zhang$^{\ddag}$, \\ {\bf \large{Liang Yang$^{\dag}$, Hongfei Lin$^{\dag}$*\thanks{* Corresponding author}}}} 

\address{$^{\dag}$School of Computer Science and Technology, Dalian University of Technology, China, \\ $^{\ddag}$School of Foreign Chinese, Dalian University of Technology, China \\
         \{dutljy, kun, 1154864382\}@mail.dlut.edu.cn\\
         \{xubo, zhangdongyu, liang, hflin\}@dlut.edu.cn\\}

\abstract{
Current methods of toxic language detection (TLD) typically rely on specific tokens to conduct decisions, which makes them suffer from lexical bias, leading to inferior performance and generalization. 
Lexical bias has both “\textit{useful}” and “\textit{misleading}” impacts on understanding toxicity. 
Unfortunately, instead of distinguishing between these impacts, current debiasing methods typically eliminate them indiscriminately, resulting in a degradation in the detection accuracy of the model.
To this end, we propose a \textbf{C}ounterfactual \textbf{C}ausal \textbf{D}ebiasing \textbf{F}ramework (\textbf{\textit{CCDF}}) to mitigate lexical bias in TLD. It preserves the “\textit{useful impact}” of lexical bias and eliminates the “\textit{misleading impact}”. 
Specifically, we first represent the total effect of the original sentence and biased tokens on decisions from a causal view.  
We then conduct counterfactual inference to exclude the direct causal effect of lexical bias from the total effect. 
Empirical evaluations demonstrate that the debiased TLD model incorporating CCDF achieves state-of-the-art performance in both accuracy and fairness compared to competitive baselines applied on several vanilla models. 
The generalization capability of our model outperforms current debiased models for out-of-distribution data. \vspace{-0.15in}
\\ \newline \textit{\textbf{Disclaimer}: The samples presented by this paper may be considered offensive or vulgar.}  %
\vspace{-0.15in}
\\ \newline  
\vspace{0.1in}
}

\begin{document}

\maketitleabstract

\section{Introduction}
\begin{figure*}[htpb]
    \centering
    \includegraphics[width=16cm]{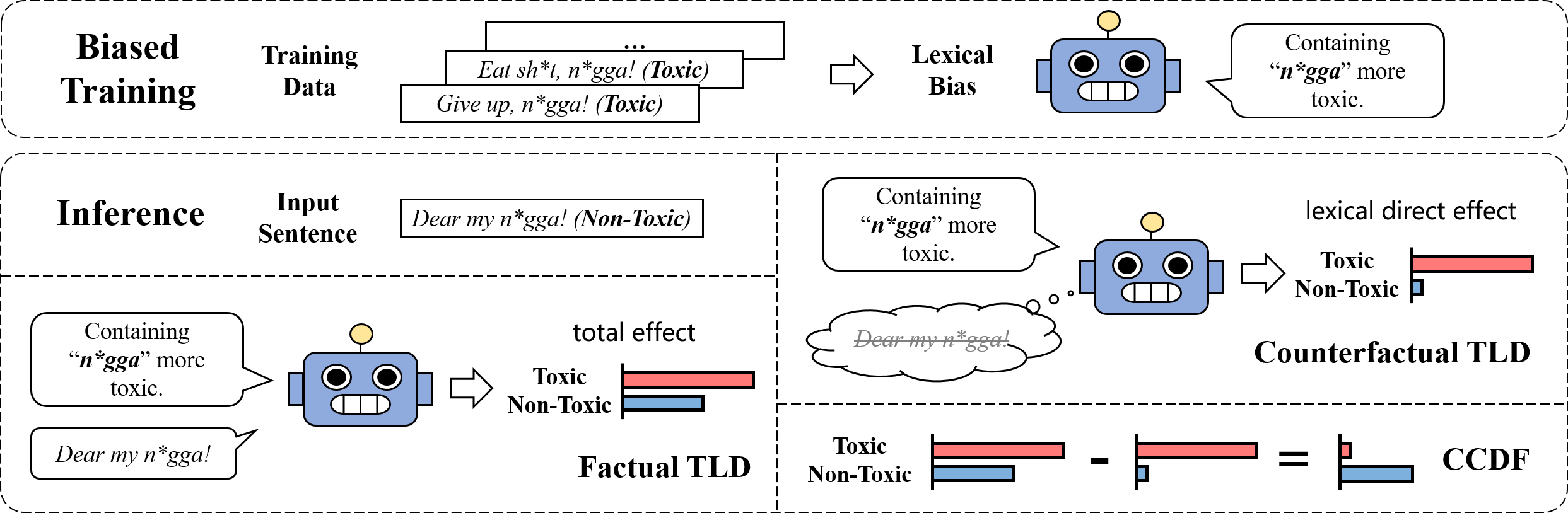}
    \vspace{-0.325in}
    \caption{ Due to the biased training, the TLD model is prone to identify all samples containing biased tokens, such as "\textit{n*gga}", as toxic language. In this paper, we present a Counterfactual Causal Debiasing Framework to mitigate lexical bias by excluding the direct causal effect of biased tokens on model decisions from the total effect. 
}
    \label{fig:introduction}
    \vspace{-0.1in}
\end{figure*}

In recent years, researchers have introduced natural language processing techniques to detect toxic language. However, due to biased training, current toxic language detection (TLD) methods are prone to relying on lexical bias to perform decisions. The lexical bias associates toxicity with the presence of biased tokens (\textit{e.g.}, identity mentions, insults, and markers of African American English) \cite{davidson2019racial, DBLP:conf/acl/ZhangBZBZZ20}, which undermines the fairness of minorities \cite{thiago2021fighting, DBLP:conf/acl/HutchinsonPDWZD20}. As an example, as shown in Figure \ref{fig:introduction}, the TLD model tends to classify all samples containing "\textit{n*gga}" (a cordial phrase for dialogue between Africans) as toxic language, due to its frequent occurrence in toxic samples during training. This actually compromises the freedom of expression of Africans \cite{DBLP:conf/acl/SapCGCS19}.
Meanwhile, lexical bias also affects the generalization ability of the TLD model, resulting in limited detection performance of the model for out-of-distribution (OOD) data \cite{vidgen2019challenges, DBLP:conf/naacl/RamponiT22, DBLP:conf/eacl/ZhouSSCS21}.

Researchers have presented several methods to mitigate lexical bias in TLD. 
Due to the expensive labor costs of constructing unbiased datasets \cite{DBLP:conf/emnlp/DinanHCW19}, many studies have attempted to weaken lexical prior while training with original data, and enable models to make decisions without the impact of the bias \cite{DBLP:conf/emnlp/SwayamdiptaSLWH20, chuang2021mitigating, DBLP:conf/naacl/RamponiT22}. 
However, these methods fail to distinguish the “\textit{useful impact}” and “\textit{misleading impact}” of lexical bias for understanding toxicity. 
In fact, lexical bias has positive effects on TLD, which was viewed as an effective surface feature for identifying toxic language in earlier work \cite{DBLP:journals/mt/Abney14, DBLP:conf/ijcai/DinakarPL15}.
As shown in Table \ref{proportion}, biased tokens are used to express toxic semantics in considerable comments.
Therefore, interpreting lexical bias as a detriment to TLD and directly eliminating the bias can lead to a significant reduction in the accuracy of debiased models \cite{DBLP:conf/eacl/ZhouSSCS21}.  
To maintain detection performance while debiasing, it is necessary to examine how lexical bias influences model decisions from the dual characteristics.

\begin{table}
\renewcommand{\arraystretch}{1.1}
\center
\small
\begin{tabular}
{m{1.25cm}<{\centering}m{1.5cm}<{\centering}m{1.5cm}<{\centering}m{1.5cm}<{\centering}} 
\bottomrule 
\textbf{Token} & \textbf{Toxic} & \textbf{Non-Toxic} &  \textbf{Ratio (\%)}  \\  
\hline 
black & 244 & 76 & 76.25 \\ 
n*gga & 541 & 17 & 96.95 \\
f*ck & 878 & 46 & 95.02 \\ 
ass & 1592 & 132 & 92.34 \\ 

\toprule
\end{tabular}  
\vspace{-0.05in}
    \caption{Proportion of toxic samples containing several biased tokens in the dataset \cite{DBLP:conf/icwsm/FountaDCLBSVSK18}, which are crawled from Twitter.}
\vspace{-0.1in}
\label{proportion}
\end{table}

In this work, we propose a novel \textbf{C}ounterfactual \textbf{C}ausal \textbf{D}ebiasing \textbf{F}ramework (CCDF) to mitigate lexical bias for TLD.
We employ causal learning techniques to examine the “\textit{useful}” and “\textit{misleading}” impact of lexical bias since it is applicable to estimating the effects of variables on model decisions \cite{pearl2018book}.
We formulate the “\textit{useful impact}” of lexical bias as the causal effect of biased tokens combined with context information on decisions, while the "\textit{misleading impact}" refers to the direct causal effect of biased tokens without introducing any context.
As shown in \figurename~\ref{fig:introduction}, two scenarios are constructed to calculate the causal effect of variables in TLD.
Specifically, we design Factual TLD to estimate the total effect of biased tokens and the input sentence on detection, which jointly influence the predicate logit of the model. 
The Counterfactual TLD is proposed to estimate the direct causal effect of biased tokens, where the model is invariant to the changes of the sentence and only relies on the lexical bias to make decisions.   
We then conduct counterfactual inference to exclude the direct causal effect of biased tokens from the total effect, thus preserving the positive effects of lexical bias and mitigating the negative effects.  

We evaluate the performance of the debiased TLD model incorporating CCDF for in-distribution data and out-of-distribution data. The experimental results demonstrate that the debiased model achieves state-of-the-art in both accuracy and fairness on several vanillas. And its migration ability outperforms current models. We further discuss the rationales of CCDF with empirical experiments.

The main contributions of this work are summarized as follows:

\begin{itemize}
    \vspace{-0.025in}
    \item We examine the positive and negative effects of lexical bias on model decisions in toxic language detection from the causal view.
    \vspace{-0.075in}    
    \item We present a Counterfactual Causal Debiasing Framework to retain the positive effects of lexical bias and mitigate negative effects, improving fairness while maintaining accuracy.  
    \vspace{-0.225in}
    \item We perform an empirical evaluation and demonstrate the effectiveness of our proposed framework in both in-distribution and out-of-distribution data\footnote{Codes of this paper are available at \url{https://github.com/DUT-lujunyu/Debias}}.
\end{itemize}

\section{Related Work}

\subsection{Debiasing for Toxic Language Detection}
Toxic language is viewed as a rude, disrespectful, or unreasonable comment that is likely to make someone leave a discussion \cite{DBLP:conf/aies/DixonLSTV18}. 
In recent years, researchers have tackled the problem of toxic language detection (TLD) using techniques of natural language processing \cite{alkhamissi-etal-2022-token, DBLP:conf/acl/TekirougluCG20, zhou2021hate, DBLP:conf/aaai/MathewSYBG021, DBLP:journals/corr/abs-2010-12472, Detoxify, DBLP:journals/inffus/MinLLZLYX23, DBLP:journals/taslp/LuLZLZZMX23, DBLP:conf/acl/LuXZMYL23}. 
Despite excellent performance on specific datasets, however, these methods over-rely on lexical bias in decision making, 
resulting in harm to the fairness of minority groups \cite{thiago2021fighting}. 
To mitigate the bias in TLD, many research efforts have been presented. The straightforward method is to balance the biased data, including using adversarial data \cite{DBLP:conf/acl-socialnlp/XiaFT20, DBLP:conf/aies/DixonLSTV18}, filtering \cite{DBLP:conf/icml/BrasSBZPSC20, DBLP:conf/emnlp/SwayamdiptaSLWH20}, relabeling \cite{DBLP:conf/eacl/ZhouSSCS21} and counterfactual data augmentation \cite{DBLP:conf/emnlp/SenSFWA21, DBLP:conf/naacl/SenS0A22}. However, the application of these methods is challenging due to the significant costs of human annotation and the uncontrollability of data selection. 

In view of this, several debiasing methods that weaken the influence of lexical priors have been presented, which can be applied directly to the original data.
\cite{DBLP:conf/acl/KennedyJDDR20, DBLP:conf/acl/AttanasioNHB22} calculated additional penalty loss for samples containing biased tokens to mitigate lexical bias. 
InvRat \cite{chuang2021mitigating} aimed to maintain invariant predictions, regardless of whether the model determines the sample contains biased tokens, thereby removing the bias.
\citet{DBLP:conf/www/Badjatiya0V19} and \citet{DBLP:conf/naacl/RamponiT22} remove and mask biased tokens directly on the original sample during the training phase, respectively. 
Motivated by LMixin \cite{DBLP:conf/emnlp/SwayamdiptaSLWH20}, \citet{DBLP:conf/eacl/ZhouSSCS21} designed a separate branch which only makes decisions based on lexical bias while training, and directly excluded it in the test phase. 
While these methods have a certain degree of debiasing effect, they fail to leverage the positive effect of lexical bias, resulting in a decrease in the accuracy of debiased TLD models. 
In contrast, our CCDF performs counterfactual reference to ensure the detection performance of the model while mitigating the lexical bias. 

\subsection{Debiasing for Other NLU Tasks}

In natural language understanding (NLU) tasks, some studies have focused on mitigating social biases in pre-trained language models to improve the fairness of models \cite{DBLP:conf/acl/ZmigrodMWC19, DBLP:conf/acl/LiangLZLSM20, DBLP:conf/iclr/ChengHYSC21, DBLP:conf/acl/GarimellaAKYNCS21, DBLP:conf/eacl/KanekoB21a, DBLP:conf/acl/GuoYA22, DBLP:conf/emnlp/HeXFC22, DBLP:journals/corr/abs-2010-06032}.
However, these methods are not applicable to debiasing for TLD due to the substantial difference in purpose. Specifically, they aim to eliminate the unbalanced model behaviors on socially sensitive topics, such as the spurious correlations between gender and careers, while the purpose of debiasing for TLD is to mitigate the dependence of model decisions on lexical bias. 
In addition, the sentences in TLD datasets are crawled from online platforms and contain more flexible word variants compared with the samples in ordinary NLU datasets \cite{DBLP:conf/cscw/0002CTS14}, which also brings challenges for debiasing \cite{zhou2021hate}.  

\section{Preliminaries}

\subsection{Causal Learning}\label{intro_causal} 
Causal learning aims to estimate the impact of variables on model decisions, which has been widely applied in various fields  \cite{DBLP:conf/cvpr/NiuTZL0W21, DBLP:conf/cvpr/TangNHSZ20, DBLP:conf/aaai/Tian0ZX22, DBLP:conf/aaai/ChoiJHH22, DBLP:conf/acl/0003FWMX20}. Here we introduce the basic concepts of causal learning. For distinction, we use the uppercase letter to denote the variable (\textit{e.g.}, \textit{X}) and the lowercase refers to its observed value (\textit{e.g.}, \textit{x}), while the lowercase letter with "*" represents the counterfactual value (\textit{e.g.}, \textit{$x^*$}). 

\begin{figure}
    \centering
    \includegraphics[width=7.5cm]{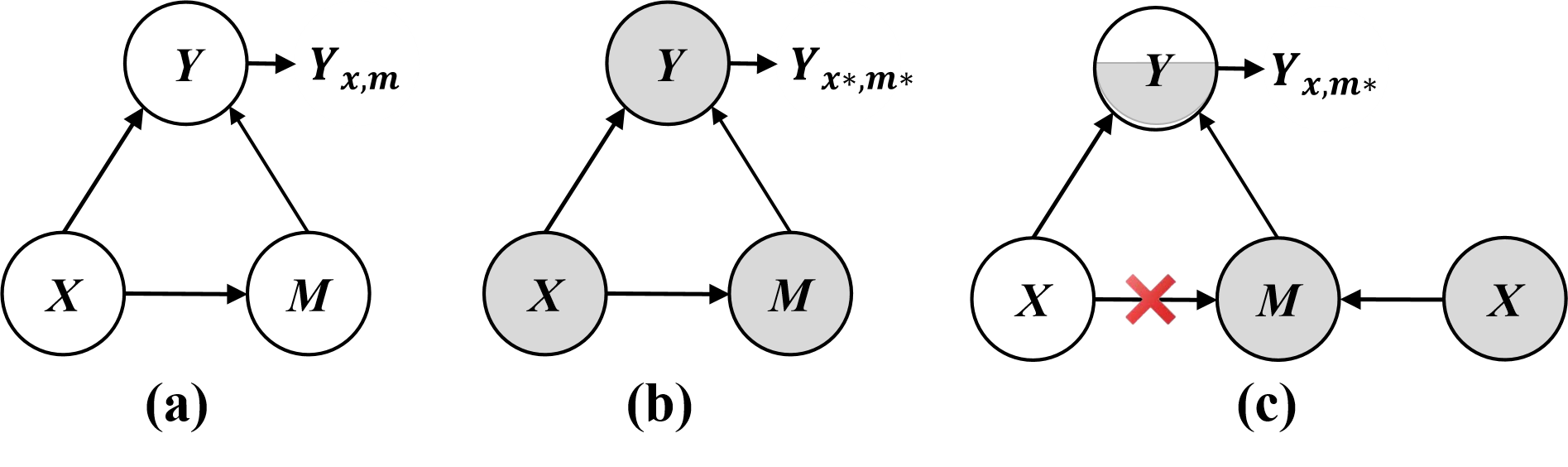}
    \vspace{-0.05in}
    \caption{Illustration of causal graph. (a) Factual scenario; (b, c) Counterfactual scenario. Where white nodes denote variables with observed values and gray nodes denote variables with counterfactual values
    }
    \vspace{-0.05in}
    \label{fig:example_graph}
\end{figure}

\textbf{Causal graph} is a directed acyclic graph $\mathcal{G} = \left \{ \mathcal{V}, \mathcal{E} \right \} $, where $\mathcal{V}$ represents the set of variables and $\mathcal{E}$ refers to the set of causal relationships from independent variables to dependent variables. An example of the causal graph is shown in \figurename~\ref{fig:example_graph}(a), which has three variables, \textit{X}, \textit{M}, and \textit{Y}. In this case, \textit{M} is a mediator between \textit{X} and \textit{Y}. Meanwhile, \textit{X} has a direct causal effect and an indirect causal effect on \textit{Y}, \textit{i.e.}, $X \rightarrow Y$ and $X\rightarrow M \rightarrow Y$. Therefore, \textit{Y} can be denoted as $Y_{x, m} = Y(X=x, M=m)$, where $m = M(X = x)$ in the factual scenario.

\textbf{Causal effects} reflect a comparison between the potential outcomes of the same individual in either treatment or not. Here we take variable \textit{X} as an example. In the factual scenario, \textit{X} is under treatment condition and gets observed value. And \textit{M} and \textit{Y} can respond to the variations of \textit{X}. In the counterfactual scenario, \textit{X} is under no-treatment condition and cannot directly affect its successor nodes. 
Furthermore, for a given variable in the causal graph, Total Effect (TE) refers to the sum of its predecessors' causal effect on it. By comparing \figurename~\ref{fig:example_graph}(a) and \figurename~\ref{fig:example_graph}(b), the TE on \textit{Y} can be written as follows:

\begin{equation}
TE = Y_{x,m}-Y_{x^*,m^*} ,
\end{equation}
where $Y_{x^*, m^*} =Y(X=x^*, M=M(X = x^*))$.
 
In causal learning, TE consists of the natural direct effect (NDE) and total indirect effect (TIE).
Where NDE estimates the direct causal effect of \textit{X} on \textit{Y} by blocking \textit{M}, resulting in \textit{M} failing to respond to the variations of \textit{X}, as shown in the comparison between \figurename~\ref{fig:example_graph}(c) and \figurename~\ref{fig:example_graph}(b). NDE can be written as follows:

\begin{equation}
NDE = Y_{x, m^*}-Y_{x^*,m^*} .
\end{equation}
Then TIE is the remaining effect of $X$ and $M$ on $Y$ after excluding the direct causal effect of $X$ on $Y$, which can be obtained by comparing TE and NDE, denoted as:

\begin{equation}
T I E=T E-N D E=Y_{x, m}-Y_{x, m^*}
\end{equation}


\begin{figure*}[htpb]
    \centering
    \includegraphics[width=16cm]{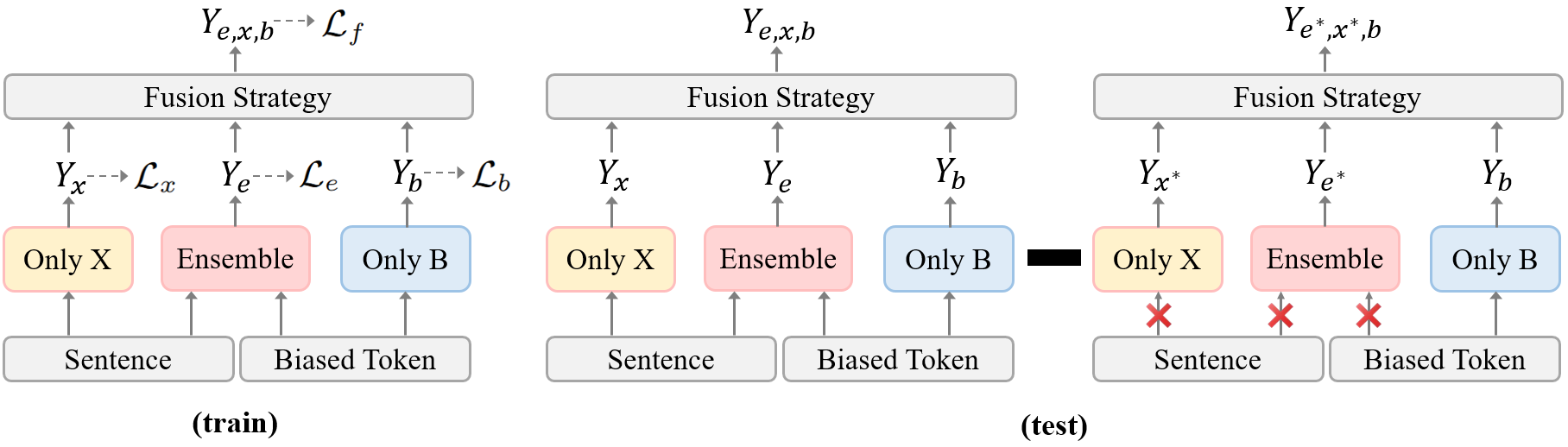}
    \vspace{-0.3in}
    \caption{The model diagram of CCDF, where \textit{Only X}, \textit{Only B} and \textit{Ensemble} represent different branch models, \textit{i.e.} $\mathcal{F}_E$, $\mathcal{F}_X$, and $\mathcal{F}_B$, respectively. The vector representations of the original sentence and biased tokens are obtained by the same encoder. $\mathcal{L}_f$, $\mathcal{L}_x$, $\mathcal{L}_e$, and $\mathcal{L}_b$ respectively refer to the loss values between each predicate logit (\textit{i.e.} $Y_{e,x,b}$, $Y_x$, $Y_e$, and $Y_b$) and the ground-truth label.}
    \vspace{-0.1in}
    \label{fig:frame}
\end{figure*}

\begin{figure}
    \centering
    \includegraphics[width=7.5cm]{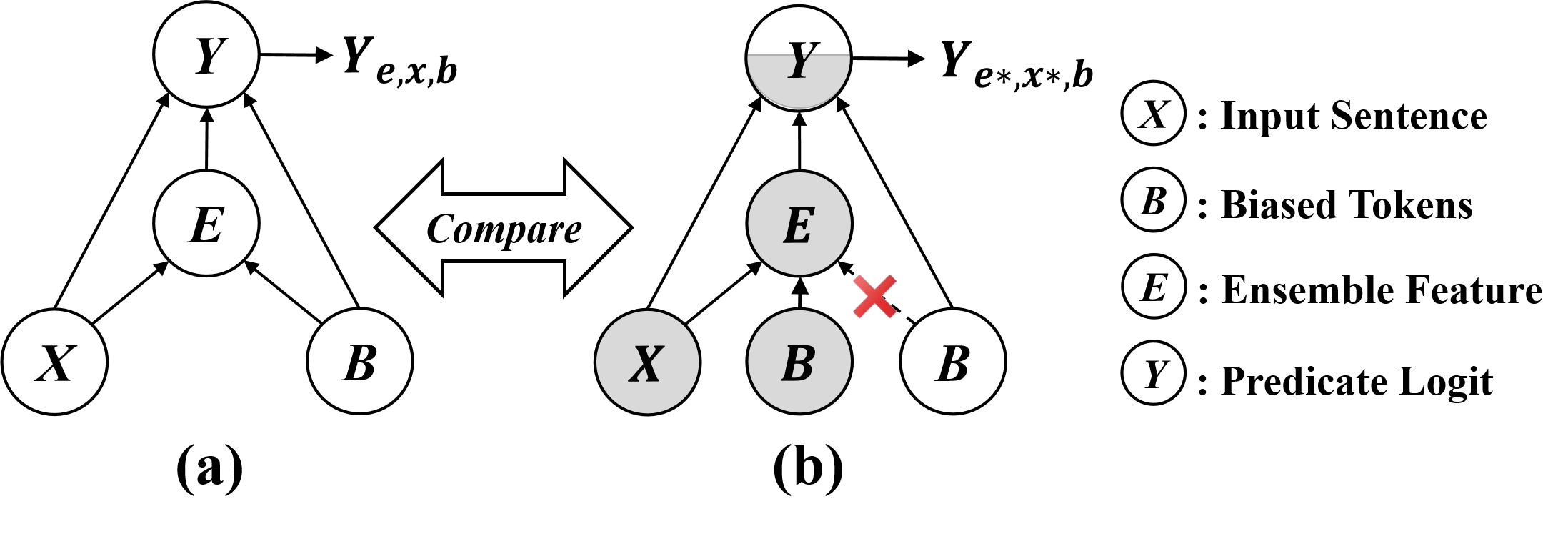}
  \vspace{-0.2in}
    \caption{Comparison between (a) Factual TLD and (b) Counterfactual TLD using causal graph.}
    \vspace{-0.1in}
    \label{fig:casual}
\end{figure}

\subsection{Problem Formulation}
Let $X = \{w_1, ..., w_n\}$ a sentence containing $n$ tokens. These tokens consist of both biased tokens, \textit{e.g.} identity mentions, denoted as $B = \{b_1, ..., b_m\}$, and unbiased tokens. To recognize the biased tokens, a public toxic lexicon \textsc{ToxTrig}\footnote{\url{https://github.com/XuhuiZhou/Toxic_Debias/blob/master/data/word_based_bias_list.csv}} \cite{DBLP:conf/eacl/ZhouSSCS21} is introduced. For an \textit{X} with a ground-truth label $y \in \{0, 1\}$, the TLD models aim to predict whether \textit{X} is toxic or non-toxic, where the prediction is denoted as \textit{Y}. 

\section{Methodology}


\subsection{Overview}

We first introduce our Counterfactual Causal Debiasing Framework (CCDF) from a causal view, and analyze the total effect of biased tokens and original sentence on model decisions during the biased training (\textit{\textit{i.e.}} Factual TLD). 
Counterfactual reference is then performed in the test phase to make debiased predictions by excluding the direct causal effect of lexical bias from the total effect (\textit{\textit{i.e.}} Counterfactual TLD). 
The diagram of our CCDF is presented in \figurename~\ref{fig:frame}.


\subsection{Causal View of CCDF}

In the CCDF, we first propose an ensemble feature \textit{E} integrating the original sentence \textit{X} and biased tokens \textit{B}. This facilitates the model to more adequately employ contextual information to determine whether biased tokens are used to express toxic semantics, maximizing the positive effects of lexical bias on model decisions. 
Then, several separate branch models are respectively utilized to obtain the logits for the three variables, \textit{\textit{i.e.}}, \textit{E}, \textit{X}, and \textit{B}.
We further incorporate these logits with a fusion function to generate the final predictions.  
The causal graph of CCDF is shown in \figurename~\ref{fig:casual} and its details are as follows. 

\textbf{Node \textit{X and B}}. These two nodes denote the original sentence \textit{X} and biased tokens \textit{B}, respectively. We employ the same encoder to obtain the vector representations of the two inputs. The corresponding separator (\textit{e.g.}, "[SEP]" in BERT \cite{DBLP:conf/naacl/DevlinCLT19}) is utilized to separate each $b_i$.

\textbf{Node \textit{E}}. It refers to the ensemble feature of \textit{X} and \textit{B}. As a mediator from \textit{X} and \textit{B} to \textit{Y}, \textit{E} can be written as follows:

\begin{equation}
E_{x,b} = E(X = x, B = b)
\end{equation}
In this work, we employ Cross Attention to integrate \textit{X} and \textit{B} to obtain \textit{E}:

\begin{equation}
E = softmax(\mathbb{X}^T \cdot \mathbb{B}) \mathbb{X},
\end{equation}
where $\mathbb{X}$ and $\mathbb{B}$ refer to the vector representations of \textit{X} and \textit{B}, respectively.


\textbf{Link \textit{X $\rightarrow$ E and B $\rightarrow$ E}}. Both \textit{X} and \textit{B} have direct causal effects on \textit{E} due to \textit{E} being built with the information of \textit{X} and \textit{B}. 

\textbf{Link \textit{E $\rightarrow$ Y, X $\rightarrow$ Y, and B $\rightarrow$ Y}}. These links denote the process by which each branch model outputs the predicate logit separately during the biased training phase. Therefore, \textit{E}, \textit{X}, and \textit{B} have direct causal effects on \textit{Y}. The branch models are represented as $\mathcal{F}_E$, $\mathcal{F}_X$, and $\mathcal{F}_B$, respectively.   

\textbf{Node \textit{Y}}. It refers to the final prediction result of CCDF, which integrates the outputs of three branch models with a fusion function. 
In the scenario of Factual TLD, all the input variables get observed values. Therefore, branch models can respond to the variation of \textit{E}, \textit{X}, and \textit{B}. And \textit{Y} can be written as follows:

\begin{equation}
Y_{e,x,b}=Y(E=e, X=x, B=b),
\end{equation}
where $e = E_{x, b}$ integrates the information of both lexical bias and context information. 

\subsection{Debiasing Inference with Casual Effect}

As the definition of total effect (TE) shown in Section \ref{intro_causal}, we compare the Factual TLD and no-treatment condition to get TE of \textit{E}, \textit{X}, and \textit{B} on \textit{Y}, which can be written as:    

\begin{equation}
TE = Y_{e,x,b}-Y_{e^*,x^*,b^*},
\end{equation}
where $e^*$, $x^*$, and $b^*$ denote the corresponding variables under no-treatment condition. 

Furthermore, based on the casual graph, the effect of biased tokens \textit{B} on the predicate logits \textit{Y} can be divided into two parts: the direct causal effect via $B \rightarrow Y$ and the indirect causal effect via $B \rightarrow E \rightarrow Y$ which incorporates the context information. Due to the significance of maintaining detection performance while debiasing, it is necessary to address the effect of \textit{B} from both sides. To mitigate the negative effects of lexical bias, the direct causal effect of \textit{B} on \textit{Y}, \textit{i.e.}, $B \rightarrow Y$, has to be eliminated from the total effect. Meanwhile, to exert the positive effects of the bias, the indirect causal effect, \textit{i.e.}, $B \rightarrow E \rightarrow Y$, should be reserved. The scenario of Counterfactual TLD is designed to estimate the direct causal effect of \textit{B}, and counterfactual inference is then conducted. Specifically, we block the direct causal effect of \textit{E} and \textit{X} on \textit{Y}, causing the branch model $\mathcal{F}_E$ and $\mathcal{F}_X$ invariant, which cannot respond to the variation of input variables \textit{E} and \textit{X}. This leads to the TLD model only relying on the lexical bias to make decisions, which is the natural direct effect (NDE) of \textit{B} on \textit{Y}.

\begin{equation}
NDE = Y_{e^*, x^*, b}-Y_{e^*,x^*,b^*} .
\end{equation}
Then the total indirect effect (TIE) of variables on \textit{Y} can be calculated by excluding NDE from TE:

\begin{equation}
T I E=T E-N D E=Y_{e,x,b}-Y_{e^*, x^*, b}.
\end{equation}
And we use TIE as the debiased prediction.

\subsection{Other implementation details}

In the implementation, we employ three separate MLPs as branch models. As shown in \figurename~\ref{fig:frame}, $\mathcal{F}_X$, $\mathcal{F}_E$, and $\mathcal{F}_B$ are running in the Factual TLD, while $\mathcal{F}_X$ and $\mathcal{F}_E$ are blocked in the Counterfactual TLD. 
Therefore, the output of each branch model can be defined as follows:

\begin{equation}
Y_b = y_b = \mathcal{F}_B(b), 
\end{equation}

\begin{equation}
Y_x= \begin{cases}
y_x=\mathcal{F}_X(x) & \text { if } X=x \\ y_x^*=c_x & \text { if } X=\varnothing  \end{cases} ,
\end{equation}

\begin{equation}
Y_e=\begin{cases}
y_e=\mathcal{F}_E(x, b) & \text { if } X=x \\
y_e^*=c_e & \text { if } X=\varnothing 
\end{cases} ,
\end{equation}
where $c_x$ and $c_e$ refer to the invariant responses of $\mathcal{F}_X$ and $\mathcal{F}_E$, respectively, which can be trained or set as hyperparameters. And $\varnothing$ denotes the no-treatment condition. 

To obtain the final predicate logit, we utilize the harmonic function to integrate $Y_e$, $Y_x$, and $Y_b$. The fused score $Y_{e,x,b}$ is as follows:

\begin{equation}
Y_{e,x,b} = h\left(Y_e, Y_x, Y_b\right)=\log \frac{Z_{e,x,b}}{1+Z_{e,x,b}}, 
\end{equation}
where $Z_{e,x,b}=tanh\left(Y_e\right) \cdot tanh\left(Y_x\right) \cdot tanh\left(Y_b\right)$ .

In the training phase, we utilize cross-entropy to calculate the difference between the predicate logits, including the output of each branch (\textit{\textit{i.e.}} $Y_e$, $Y_x$, and $Y_b$) and the fused score $Y_{e,x,b}$, and the ground-truth label $y$. The final loss function is defined as follows: 

\begin{align}
\mathcal{L}_{all}
& = \mathcal{L}_f + \mathcal{L}_e + \mathcal{L}_x + \mathcal{L}_b  \\ 
\nonumber & = \mathcal{L}(Y_{e,x,b}, y)+\mathcal{L}(Y_e, y)+\mathcal{L}(Y_x, y)+\mathcal{L}(Y_b, y).
\end{align}
The parameters of TLD models are optimized by minimizing $\mathcal{L}_{all}$. 
In addition, since \textit{X} and \textit{B} share the same encoder, we do not backpropagate $\mathcal{L}(Y_b, y)$ to the encoder, preventing the encoder from learning lexical bias directly. And $\mathcal{L}(Y_b, y)$ is only used to update parameters of $\mathcal{F}_B$.

\section{Experiments}

\subsection{Datasets and Evaluation Metrics}

For fair comparisons with baselines of debiasing methods for the TLD model, we followed \citet{DBLP:conf/eacl/ZhouSSCS21} and selected the same benchmarks in both in-distribution and out-of-distribution data. 
Specifically, we first conducted the main experiment on \cite{DBLP:conf/icwsm/FountaDCLBSVSK18}, which has 32K toxic and 54K non-toxic samples crawled from Twitter. Referenced by \citet{DBLP:conf/eacl/ZhouSSCS21}, we focused on three kinds of lexical biases, including non-offensive minority identity (\textsc{nOI}), \textit{e.g.}, \textit{gay}, offensive minority identity (\textsc{OI}), \textit{e.g.}, \textit{n*gga}, and offensive non-identity (\textsc{OnI}), \textit{e.g.}, \textit{f*ck}.
Overall accuracy ($Acc$) and $F_1$ are used to measure the detection performance of TLD models. Then $F_1$ and false positive rate ($FPR$) on the samples containing \textsc{nOI}, \textsc{OI}, and \textsc{OnI} are respectively reported, evaluating the degree of lexical bias in the model. Intuitively, the lower the $FPR$, the less the model relies on lexical bias in decision making, and the fairer the model. 

We then evaluated the performance of trained models on OOD data. We used the test set of \cite{DBLP:conf/emnlp/DinanHCW19} as the adversarial dataset, which contains 6k artificial sentences (including 600 toxic samples). 
The language style of these artificially constructed samples is quite different from the in-distribution data crawled from Twitter and has a more standardized character. In addition, many of the toxic samples in this dataset are implicit and do not contain insults towards minorities. This 
presents a serious challenge to the generalization capability of TLD models. 
Here we use the accuracy and weighted $F_1$ as evaluation metrics.

\begin{table*}[htbp]
\small
  \centering
    \begin{tabular}{m{1.85cm}<{\centering}|m{0.9cm}<{\centering}m{1.1cm}<{\centering}|m{0.9cm}<{\centering}m{0.8cm}<{\centering}m{0.9cm}<{\centering}m{0.9cm}<{\centering}m{0.9cm}<{\centering}m{1.1cm}<{\centering}|m{0.9cm}<{\centering}m{1.05cm}<{\centering}}    \toprule
          & \multicolumn{2}{c|}{Test \ (12893)} & \multicolumn{2}{c}{nOI \ (602)} & \multicolumn{2}{c}{OI \ (553)} & \multicolumn{2}{c|}{OnI \ (3236)} & \multicolumn{2}{c}{OOD \ (6000)} \\
\cmidrule{2-11}    Method & $Acc$ $\uparrow$  & $F_1$ $\uparrow$ & $F_1$ $\uparrow$ & $FPR\downarrow$  & $F_1$ $\uparrow$ & $FPR \downarrow$  & $F_1$ $\uparrow$ & $FPR$ $\downarrow$  & $Acc$ $\uparrow$ & $F_1$ $\uparrow$\\
    \midrule
    \multicolumn{11}{l}{\textit{weakening lexical prior with BERT-base:}} \\
    \midrule 
    Vanilla† & 93.53\textsubscript{0.1} & 91.39\textsubscript{0.1} & 89.29\textsubscript{0.3} & 9.22\textsubscript{0.4} & 98.90\textsubscript{0.0} & 85.71\textsubscript{3.4} & 97.19\textsubscript{0.0} & 66.34\textsubscript{1.4} & 91.28\textsubscript{0.1} & 81.43\textsubscript{2.4} \\
    \midrule
    Masking & 93.24\textsubscript{0.1} & 91.08\textsubscript{0.1} & \textbf{89.33}\textsubscript{0.2} & 9.56\textsubscript{0.4} & \textbf{98.80}\textsubscript{0.2} & 83.33\textsubscript{3.4} & \textbf{97.24}\textsubscript{0.0} & 64.88\textsubscript{0.5} & 91.55\textsubscript{0.0} & 81.71\textsubscript{0.6} \\
    LMixin  & 91.85\textsubscript{0.5} & 89.51\textsubscript{0.5} & 87.19\textsubscript{0.2} & 10.24\textsubscript{2.7} & 98.33\textsubscript{0.0} & \textbf{74.52}\textsubscript{0.0} & 97.08\textsubscript{0.1} & 59.51\textsubscript{1.8} & 91.58\textsubscript{0.1} & 83.67\textsubscript{3.2} \\
    CCDF(ours) & \textbf{93.75}\textsubscript{0.0} & \textbf{91.59}\textsubscript{0.0} & 88.54\textsubscript{0.7} & \textbf{4.10}\textsubscript{0.9} & 98.62\textsubscript{0.8} & 78.57\textsubscript{0.0} & 97.15\textsubscript{0.2} & \textbf{59.02}\textsubscript{2.3} & \textbf{91.63}\textsubscript{0.1} & \textbf{85.81}\textsubscript{1.2} \\
    \midrule
    \multicolumn{11}{l}{\textit{weakening lexical prior with RoBERTa-base:}} \\
    \midrule
    Vanilla† & 94.04\textsubscript{0.1} & 91.70\textsubscript{0.1} & 90.10\textsubscript{0.3} & 8.40\textsubscript{0.4} & 98.60\textsubscript{0.0} & 81.00\textsubscript{3.4} & 97.00\textsubscript{0.0} & 63.40\textsubscript{1.4} & 92.19\textsubscript{0.1} & 81.78\textsubscript{2.4} \\
    \midrule
    Masking & 93.91\textsubscript{0.1} & \textbf{92.01}\textsubscript{0.1} & \textbf{89.60}\textsubscript{0.2} & 5.30\textsubscript{0.4} & \textbf{98.21}\textsubscript{0.2} & 80.95\textsubscript{3.4} & \textbf{97.32}\textsubscript{0.0} & 62.76\textsubscript{0.5} & 92.27\textsubscript{0.0} & 82.35\textsubscript{0.6} \\
    LMixin  & 92.05\textsubscript{0.5} & 90.53\textsubscript{0.5} & 87.51\textsubscript{0.2} & 6.35\textsubscript{2.7} & 97.93\textsubscript{0.0} & \textbf{71.43}\textsubscript{0.0} & 97.14\textsubscript{0.1} & 63.09\textsubscript{1.8} & 91.92\textsubscript{0.1} & 83.11\textsubscript{3.2} \\
    InvRat† & -   & 91.00\textsubscript{0.5} & 85.50\textsubscript{1.6} & 3.40\textsubscript{0.6} & 97.50\textsubscript{1.0} & 76.20\textsubscript{3.4} & 97.20\textsubscript{0.2} & 61.10\textsubscript{1.5} & -   & - \\
    CCDF(ours) & \textbf{94.05}\textsubscript{0.0} & 91.86\textsubscript{0.0} & 85.91\textsubscript{0.7} & \textbf{2.85}\textsubscript{0.9} & 97.69\textsubscript{0.8} & \textbf{71.43}\textsubscript{0.0} & 97.12\textsubscript{0.2} & \textbf{57.23}\textsubscript{3.3} & \textbf{92.39}\textsubscript{0.1} & \textbf{86.12}\textsubscript{1.2} \\
    \midrule
    \multicolumn{11}{l}{\textit{weakening lexical prior with RoBERTa-large:}} \\
    \midrule
    Vanilla‡ & 94.21\textsubscript{0.0} & 92.33\textsubscript{0.0} & 89.76\textsubscript{0.3} & 10.24\textsubscript{1.3} & 98.84\textsubscript{0.1} & 85.71\textsubscript{0.0} & 97.34\textsubscript{0.1} & 64.72\textsubscript{0.8} & 92.20\textsubscript{0.1} & 82.20\textsubscript{2.0} \\
    \midrule
    Masking & 93.67\textsubscript{0.1} & 91.75\textsubscript{0.1} & \textbf{87.56}\textsubscript{0.7} & 8.19\textsubscript{1.1} & 98.40\textsubscript{0.5} & 83.33\textsubscript{3.4} & 97.40\textsubscript{0.1} & 61.79\textsubscript{2.3} & 91.93\textsubscript{0.2} & 84.01\textsubscript{2.2} \\
    LMixin‡ & 90.44\textsubscript{0.7} & 86.94\textsubscript{1.1} & 85.47\textsubscript{0.3} & 11.15\textsubscript{1.7} & 97.64\textsubscript{0.3} & \textbf{71.43}\textsubscript{0.0} & 90.41\textsubscript{1.8} & \textbf{44.55}\textsubscript{1.5} & -   & - \\
    LMixin & 91.67\textsubscript{1.1} & 89.58\textsubscript{1.1} & 86.76\textsubscript{0.8} & 6.94\textsubscript{0.7} & 98.12\textsubscript{0.3} & 78.57\textsubscript{2.9} & 96.95\textsubscript{0.1} & 56.10\textsubscript{1.2} & 91.95\textsubscript{0.1} & 85.35\textsubscript{1.9} \\
    CCDF(ours) & \textbf{94.15}\textsubscript{0.1} & \textbf{92.07}\textsubscript{0.1} & 86.65\textsubscript{0.9} & \textbf{3.75}\textsubscript{1.0} & \textbf{98.49}\textsubscript{0.3} & 78.57\textsubscript{0.0} & \textbf{97.42}\textsubscript{0.1} & 58.54\textsubscript{3.2} & \textbf{92.33}\textsubscript{0.0} & \textbf{86.40}\textsubscript{1.6} \\
    \midrule
    \midrule
    \multicolumn{11}{l}{\textit{balancing training data with RoBERTa-large:}} \\
    \midrule
    AFLite & 93.86\textsubscript{0.1} & 91.94\textsubscript{0.1} & 90.21\textsubscript{0.4} & 8.22\textsubscript{1.1} & 98.90\textsubscript{0.0} & 85.71\textsubscript{0.0} & 97.32\textsubscript{0.1} & 62.44\textsubscript{0.0} & 91.34\textsubscript{0.2} & 79.61\textsubscript{2.3} \\
    \ \ \ \ \ \ \ w/ CCDF & 93.85\textsubscript{0.1} & 91.83\textsubscript{0.0} & 86.36\textsubscript{0.6} & 3.83\textsubscript{0.7} & 98.78\textsubscript{0.2} & 78.57\textsubscript{0.0} & 97.31\textsubscript{0.1} & 59.35\textsubscript{2.8} & 91.73\textsubscript{0.1} & 82.56\textsubscript{1.9} \\
    \midrule
    DataMaps & 94.33\textsubscript{0.1} & 92.45\textsubscript{0.1} & 89.16\textsubscript{0.7} & 7.39\textsubscript{1.0} & 98.87\textsubscript{0.1} & 85.71\textsubscript{0.0} & 97.54\textsubscript{0.0} & 64.39\textsubscript{1.4} & 91.54\textsubscript{0.3}    & 81.62\textsubscript{1.3} \\
    \ \ \ \ \ \ \ w/ CCDF & 94.25\textsubscript{0.0} & 92.20\textsubscript{0.1} & 88.11\textsubscript{0.4} & 3.75\textsubscript{0.9} & 98.34\textsubscript{0.1} & 78.57\textsubscript{0.0} & 97.13\textsubscript{0.1} & 60.49\textsubscript{1.1} & 92.20\textsubscript{0.3}    & 83.64\textsubscript{1.8} \\
    \bottomrule
    \end{tabular}%
    \vspace{-0.025in}
  \caption{Evaluation on the test set of \cite{DBLP:conf/icwsm/FountaDCLBSVSK18} and adversarial dataset \cite{DBLP:conf/emnlp/DinanHCW19}. Results show the mean and s.d. (subscript) of $Acc$ and $F_1$ across 3 runs, as well as $F_1$ and $FPR$ towards test samples containing specific mentions in \textsc{ToxTrig}, including \textsc{nOI}, \textsc{OI}, and \textsc{OnI}. 
  The \textbf{best} results of debiasing methods that weaken lexical prior are highlighted in each column. 
  † : results reported in \citet{chuang2021mitigating}; ‡: results reported in \citet{DBLP:conf/eacl/ZhouSSCS21}. $\uparrow$: greater the better; $\downarrow$: lower the better.}
  \vspace{-0.05in}
  \label{main_res}%
\end{table*}%

\subsection{Baselines and Experimental Settings}
We conducted various baselines to mitigate lexical bias in TLD models, including both weakening lexical prior with original data and unbiased training with data filtering. 
For methods of weakening lexical prior, we selected Masking \cite{DBLP:conf/naacl/RamponiT22}, LMixin \cite{DBLP:conf/emnlp/SwayamdiptaSLWH20} and InvRat \cite{chuang2021mitigating}. We evaluated the methods on three commonly used PLMs, including BERT-base \cite{DBLP:conf/naacl/DevlinCLT19}, RoBERTa-base and RoBERTa-large \cite{DBLP:journals/corr/abs-1907-11692}.
For methods of data filtering, two data filtering methods were introduced and applied on RoBERTa-large, including AFLite \cite{DBLP:conf/icml/BrasSBZPSC20} and DataMaps \cite{DBLP:conf/emnlp/SwayamdiptaSLWH20}. The filtered training data size is 33\% of the original training set. 

In the experimental stage, we trained TLD models on the training set of \cite{DBLP:conf/icwsm/FountaDCLBSVSK18} and saved their best parameters on the validation set. Then we respectively evaluated the performance of models on the test set and adversarial dataset.
To further prove the generalization of our CCDF, we also evaluated its performance on balanced training data filtered by AFLite and DataMaps, respectively.
We did not perform any pre-processing of the datasets, or any hyperparameter search, but followed all the settings in \citet{DBLP:conf/eacl/ZhouSSCS21}. We use one NVIDIA GeForce RTX 3090 to perform the experiments. AdamW is used as the optimizer for model training. 


\subsection{Quantitative Results}

\subsubsection{Main Discussions}

\tablename~\ref{main_res} shows our empirical evaluation results of both in-distribution and OOD data. And we have the following findings:

\textbf{RQ1: Performance of Weakening Lexical Prior Methods on In-distribution Data.} 
Overall, whereas most debiasing methods of weakening the lexical prior can improve the fairness of TLD models, they also lead to a reduction in the accuracy of models' detection for in-distribution data.  
Here we take LMixin as an example, which is a competitive baseline for mitigating lexical bias. After the adoption of LMixin, the accuracy and $F_1$ value of the models decrease on average by almost 2.5\% for the samples of the test set.  

We also notice that models introducing Masking have little performance degradation on in-distribution data and have almost the worst ability to mitigate the bias. This is because Masking itself is an incomplete method of debiasing, and the biased tokens are only masked in the training set and remain in the validation and test sets. Therefore, even though it prevents the model from learning the lexical bias during the training phase, the model can still make decisions on the test set based on the lexical prior learned during pre-training. Since the PLM itself has already learned the semantics of these tokens, the overall impact of Masking on model performance is minimal.

In contrast, with the introduction of our CCDF, TLD models outperform the original vanilla in detecting in-distribution data, and achieve state-of-the-art debiasing effects for the lexical bias of nOI and OI.
This demonstrates that our method can effectively mitigate bias while preserving model detection performance, achieving a trade-off between model accuracy and fairness.

\textbf{RQ2: Performance of Weakening Lexical Prior Methods on Out-of-distribution Data.} 
Compared to the original vanilla, TLD models introducing the debiasing method exhibit a better detection performance on OOD data. This indicates that removing spurious associations between biased tokens and labels can improve the generalization ability of the models. 
Moreover, our CCDF significantly outperforms other debiasing methods to provide maximum benefit to the model, with an improvement of approximately 4.5\% in the $F_1$ value.
Meanwhile, We also find that the accuracy of TLD models in detecting OOD data is much higher than the $F_1$ value. This is because the adversarial dataset contains implicitly toxic samples that the models often misclassify as non-toxic, leading to a decrease in $F_1$.
We will further conduct an error analysis to illustrate these samples in Section \ref{qua_ana} below.


\textbf{RQ3: Comparison with Debiasing Methods of Data Filtering.} 
Models trained on the balanced data have high performance on in-distribution data.
This is because data filtering methods are able to select the most efficient samples, enabling models to achieve optimal results while utilizing minimal data.  
However, the debiasing effect of these models is limited, especially for the lexical bias of OI and OnI. 
This is due to the fact that data filtering only relies on the confidence probability of the model on the sample without effectively eliminating samples containing lexical bias.
Furthermore, for OOD data, we reach the same conclusion as \citet{zhou2021hate} that data filtering methods have poor generalization ability compared to debiasing methods that weaken lexical prior due to insufficient training data (only 33\% of the original data after filtering). 
Besides, these methods require a relatively high time overhead to perform additional training rounds for data selection.
In contrast, our method is more efficient by training on the original dataset and has a better effect of mitigating bias. 

Furthermore, we evaluate the performance of CCDF on balanced training data. 
The results indicate that the incorporation of CCDF can further enhance the fairness of the model trained on the balanced data, resulting in a significantly lower $FPR$. 
Meanwhile, the debiased model retains the benefits of the filtering data method, maintaining detection performance on in-distribution data. The results of OOD data demonstrate its enhanced capacity for generalization.

\begin{table}
\small
  \centering
    \begin{tabular}{l|c|c|c}
    \toprule
          & Test  & nOI   & OOD \\
\cmidrule{2-4}    \multicolumn{1}{c|}{Method} & $F_1$ $\uparrow$ & $FPR$ $\downarrow$  & $F_1$ $\uparrow$\\
    \midrule
    \multicolumn{1}{l|}{RoBERTa-base} & 91.70\textsubscript{0.1}     &  8.40\textsubscript{0.4} & 81.78\textsubscript{2.4} \\
    \multicolumn{1}{l|}{CCDF} & \textbf{91.86}\textsubscript{0.0}     &  \textbf{2.85}\textsubscript{0.9} & \textbf{86.12}\textsubscript{1.2} \\
    \midrule
    \ \ w/o $\mathcal{F}_e$ &  91.82\textsubscript{0.1}     &    3.57\textsubscript{1.2}   & 83.69\textsubscript{1.7} \\
    \ \ w/o $\mathcal{F}_x$  & 91.86\textsubscript{0.1}      & 3.04\textsubscript{0.8}      & 84.23\textsubscript{1.5} \\
    \ \ w/o $\mathcal{F}_x$ \& $\mathcal{F}_b$  & 91.76\textsubscript{0.1}      &  7.64\textsubscript{0.5}     & 82.24\textsubscript{1.7} \\
    \bottomrule
    \end{tabular}%
    \vspace{-0.025in}
  \caption{Ablation experiments on branch model, where w/o $\mathcal{F}_x$ \& $\mathcal{F}_b$ refers to the model which only relies on $\mathcal{F}_e$ to make decisions.}
  \vspace{-0.05in}
  \label{ablation}%
\end{table}%


\begin{table*}
\renewcommand{\arraystretch}{1.2}
\small
  \centering
    \begin{tabular}{clcccc}
    \toprule
          & Sentence & Label  & Vanilla & LMixin & CCDF \\
    \midrule
    \multirow{2}[2]{*}[3pt]{(a)} & @user @user You don’t have to pay for their \textbf{\textit{bullshit}} read your & \multirow{2}[2]{*}[3pt]{\ton} & \multirow{2}[2]{*}[3pt]{\tox} & \multirow{2}[2]{*}[3pt]{\ton} & \multirow{2}[2]{*}[3pt]{\ton} \\
          & rights read the law I don’t pay fo. . . &       &       &       &  \\
    \midrule
    (b)   & RT @user: my ex so ugly to me now like...i’ll beat that \textbf{\textit{hoe ass}}. & \tox  & \tox  & \ton  & \tox \\
    \midrule
    \multirow{2}[2]{*}[3pt]{(c)} & @user Stop that, it’s not your fault a \textbf{\textit{scumbag}} decided to steal & \multirow{2}[2]{*}[3pt]{\tox} & \multirow{2}[2]{*}[3pt]{\tox} & \multirow{2}[2]{*}[3pt]{\ton} & \multirow{2}[2]{*}[3pt]{\ton} \\
          & otems which were obviously meant for someone i. . . &       &       &       &  \\
    \midrule
    (d)   & He should go back to his lad. & \tox  & \ton  & \ton  & \ton \\
    \bottomrule
    \end{tabular}%
\caption{Examples from the test set and adversarial dataset with predictions from vanilla (RoBERTa-base), LMixin, and our CCDF. \tox~ denotes toxic label and \ton~ denotes non-toxic label. Biased tokens in examples are highlighted.}
  \vspace{-0.1in}
  \label{Qualitative}%
\end{table*}%

\subsubsection{Ablation Experiments}

Here we further conduct ablation experiments on the branch model of our CCDF. The experimental results are shown in Table \ref{ablation}. From the results, we obtain the following conclusions:

(1) Ablation of branch model $\mathcal{F}_x$ or $\mathcal{F}_e$ has little effect on the performance of the model on in-distribution data. 
This is because counterfactual reasoning can still be performed on the ablated model to remove the negative effects of lexical bias and improve fairness.
Meanwhile, the accuracy of the ablated model on OOD data is significantly reduced. This reflects that joint training of multiple branches facilitates the generalization ability of the model. 
In addition, whether ablating branch model $\mathcal{F}_x$ or $\mathcal{F}_e$, ablated CCDF has a more competitive performance than baselines, illustrating the effectiveness of our framework.

(2) We also find that CCDF without the branch model $\mathcal{F}_x$ has higher accuracy and better fairness than without $\mathcal{F}_e$, and has stronger generalization on OOD data. Meanwhile, compared with vanilla, $\mathcal{F}_e$ performs better. This reflects the fact that integrating context information and biased tokens for making decisions can fully exploit the positive effects of lexical bias, demonstrating the significance of ensemble features to TLD.

\subsection{Qualitative analysis} \label{qua_ana}
In this section, we further illustrate the capability of our CCDF in mitigating the lexical bias for TLD, providing several examples shown in \tablename~\ref{Qualitative}. We choose RoBERTa-base as vanilla and list predictions from LMixin and our CCDF for comparison.

For Exp. (a), the vanilla incorrectly predicts this non-toxic sentence as toxic, due to the biased token "\textit{bullshit}", which is itself a high-frequency swear word, even though in context it does not express the semantics of toxic. And the vanilla introducing either LMixin or CCDF can perform a debiased decision. 
For Exp. (b), we find that the vanilla and our CCDF correctly predict the label, while LMixin does not. This is because it ignores the positive effects of biased tokens (i.e., "\textit{hoe}" and "\textit{ass}"). In contrast, our CCDF integrates context information and lexical bias to preserve the positive effect and provide more accurate predictions.   

To gain more insights into the performance of our model, we list two sentences that our model misclassifies, Exp. (c) and (d) shown in \tablename~\ref{Qualitative}, and conduct an error analysis. 
For Exp. (c), both debiased models introducing LMixin and CCDF incorrectly predict this toxic sentence as non-toxic, while vanilla correctly predicts the label. This is because the toxicity degree of "\textit{scumbag}" is small, leading debiased models to consider the sample as non-toxic. 
And for Exp. (d), which implicitly expresses sarcasm towards LGBTQ, vanilla and LMixin also make incorrect predictions. This result reflects that current TLD models still lack enough world knowledge to capture potential toxicity, resulting in poor detection of samples that do not contain insults.



\section{Conclusion}
In this paper, we propose a Counterfactual Causal Debiasing Framework to mitigate lexical bias in toxic language detection. 
We formulate the bias as the causal effect of biased tokens on decisions and build a causal graph to analyze the causal relationship between variables and predicate logits.
In the training phase, our framework integrates context information to leverage the positive effects of lexical bias, guaranteeing the detection performance of the model. The negative effects of the bias are then removed from the total effect by performing counterfactual reasoning during the testing phase.
In the experiments, we show that our framework significantly outperforms the state-of-the-art debiasing methods on both accuracy and fairness for TLD. Furthermore, we demonstrate that the debiased model employing our framework has an excellent generalization capability in out-of-distribution data. In the future, we will further evaluate the debiasing effect of CCDF for other NLU tasks, as well as the performance applied to large language models.



\section*{Limitations}

In this work, we utilized a toxic lexicon as prior knowledge to recognize biased tokens. However, since lexical bias is generated during model training rather than being artificially proposed \cite{DBLP:conf/acl/HutchinsonPDWZD20}, the external lexicon may not align with the biased tokens learned by the TLD model during training.
Consequently, an incomplete lexicon could result in new lexical bias, which would negatively impact the fairness of the model \cite{DBLP:conf/acl/JoshiH22}. 
Furthermore, our study did not investigate sentence-level dialectal bias, such as African American English (AAE), which is officially considered a less appropriate language variety, and this exacerbates racial bias \cite{DBLP:conf/acl/SapCGCS19}.
Based on the above, it is imperative to acknowledge that CCDF should not be perceived as a universal solution to mitigate all the bias in TLD. Instead, it should be regarded as an innovative attempt to highlight certain aspects of a complex, elusive, and multifaceted problem.
In the future, we will investigate techniques for identifying lexical bias from the standpoint of model training and examine strategies for alleviating dialectal bias in the TLD model.

\section*{Ethics Statement}
Due to the research field of this work, we recognize that the examples provided in the paper may have a negative impact on certain minority groups. 
However, our values include honesty, integrity, respect, fairness, and responsibility. We are committed to treating all individuals with dignity and respect, and to promoting a culture of inclusivity and diversity.
Therefore, the views and conclusions presented in these examples should not be interpreted as reflecting the opinions or beliefs expressed or implied by the authors. Our hope is that the advantages of this research outweigh any potential risks.

\section*{Acknowledgment}
This research is supported by the National Natural Science Foundation of China (No. 62376051, 62076046, 62076051, 62066044), and Liaoning Province Applied Basic Research Program (No. 2022JH2/101300270).
We would like to thank all reviewers for their constructive comments.

\section*{Bibliographical References}\label{sec:reference}
\vspace{-0.2in}
\bibliographystyle{lrec-coling2024-natbib}
\bibliography{all_ref}


\newpage
\appendix
\renewcommand\thefigure{\Alph{section}\arabic{figure}}    
\renewcommand\thetable{\Alph{section}\arabic{table}}

\section{Experimental Details} \label{exp_app}

\subsection{Hyperparameter Setting}
We use one NVIDIA GeForce RTX 3090 to perform the experiments. AdamW is used as the optimizer for model training. The invariant responses of $\mathcal{F}_X$ and $\mathcal{F}_E$ in the Counterfactual TLD, i.e. $c_x$ and $c_e$, are obtained by training. Other details of hyperparameters are directly followed \cite{DBLP:conf/eacl/ZhouSSCS21}, listed in \tablename~\ref{hyp}.
While under non-optimal hyperparameters, debiased models with CCDF still obtain state-of-the-art performance in both accuracy and fairness across most datasets compared with other methods.

\setcounter{table}{0}    
\begin{table}[htpb]
\renewcommand{\arraystretch}{1.2}
\begin{small}
\centering
    \begin{tabular}{cc}
    \hline
    \textbf{Hyperparameter} & \textbf{Value} \\
    \hline
    epochs & 3    \\
    saved steps & 1000    \\
    batch size &8  \\
    learning rate &1e-5  \\
    dropout &0.1  \\
    hidden state of MLPs & 256 \\  
    padded length of sentence & 128 \\
    padded length of biased tokens & 16 \\
    \hline
    \end{tabular}
    \caption{The hyperparameters of the experiments.}
    \label{hyp}
\end{small}
\end{table}

\subsection{Baseline Introduction and Implementation}

Here we further introduce the baselines and their implementation details. The same hyperparameters are utilized as our CCDF.



\textbf{Masking} \cite{DBLP:conf/naacl/RamponiT22}: In the training set, the biased tokens are masked, while in the validation set and test set, they are still retained.

\textbf{LMixin} \cite{DBLP:conf/eacl/ZhouSSCS21}: In the training phase, model decisions depend on the outputs of two branch models, whose inputs are the original sentence and biased tokens respectively, like our CCDF without $\mathcal{F}_E$. In the test phase, the model makes predictions only based on the sentence.

\textbf{InvRat} \cite{chuang2021mitigating}: InvRat is a three-player framework consisting of an environment-agnostic predictor, an environment-aware predictor, and a rationale generator. Therefore, three independent RoBERTa models are running at the same time during the training phase.


\textbf{AFLite} \cite{DBLP:conf/icml/BrasSBZPSC20}: An ensemble of simple linear classifiers is trained and tested on the dataset. Samples that are correctly classified by most of the classifiers in the ensemble are considered to contain lexical bias and are discarded. The algorithm is iterative until the remaining data reaches the target size.

\textbf{DataMaps} \cite{DBLP:conf/emnlp/SwayamdiptaSLWH20}: For a specific model, there are different regions in a dataset, including easy, hard, and ambiguous regions. These regions are identified based on the confidence of the model in the true category of examples, and the variation of this confidence during the training phase. DataMaps-Easy, DataMaps-Ambiguous, and DataMaps-Hard subsets of the dataset are then created \cite{DBLP:conf/icwsm/FountaDCLBSVSK18}. 

Following \cite{zhou2021hate}, the size of filtered subsets is set to 33\% of the original training set for both filtering methods, and label proportions are preserved. Then a RoBERTa-large classifier is fine-tuned on filtered subsets. 

\section{Causal View of Ablated CCDF} \label{ablated_intro} 

\setcounter{figure}{0}    
\begin{figure}
    \centering
    \includegraphics[width=7.5cm]{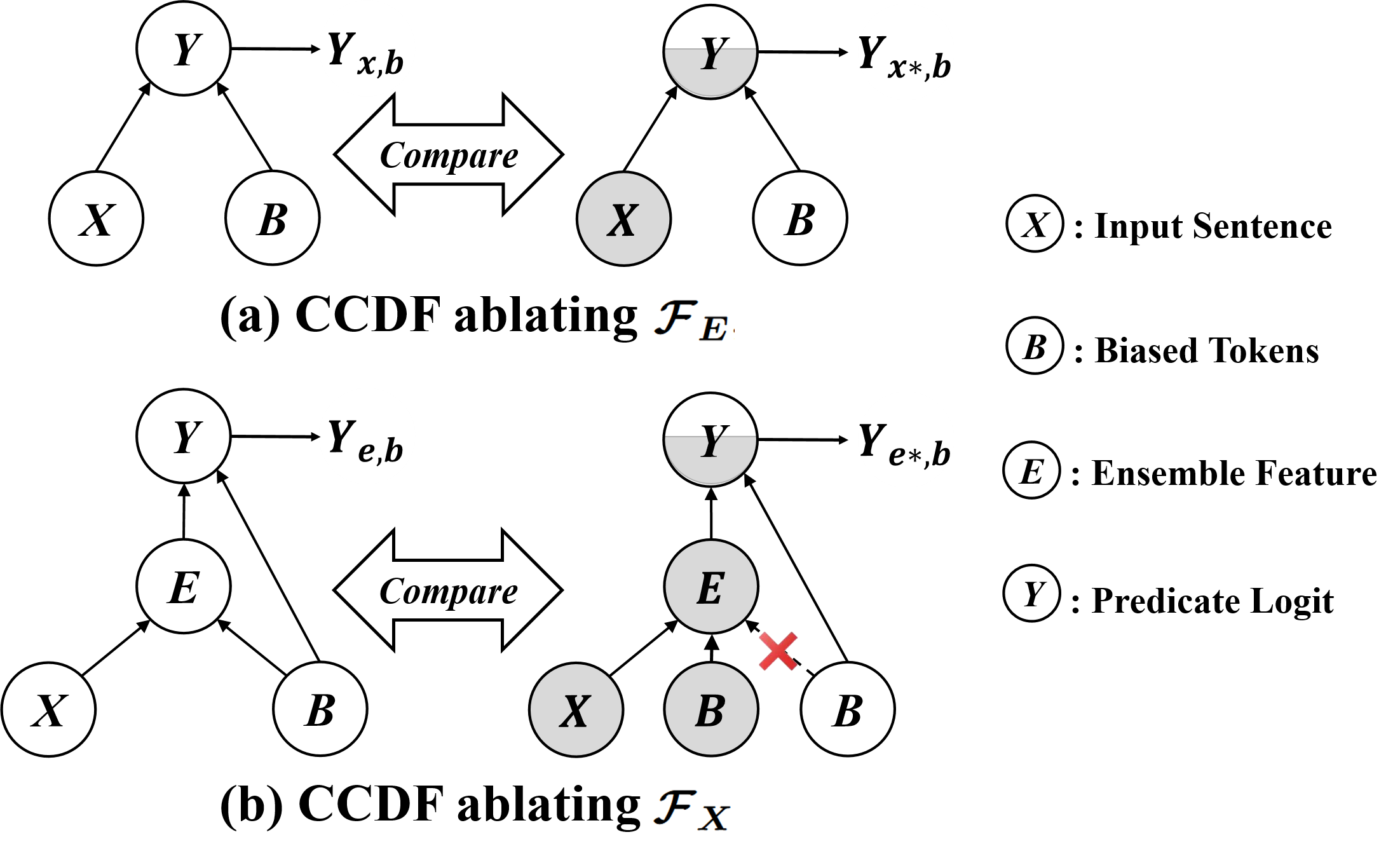}
    \caption{Causal graph of ablated CCDF.}
    \label{fig:casual_ab}
\end{figure}

Here we introduce the ablated CCDF from the causal view. As shown in \figurename~\ref{fig:casual_ab}(a), for CCDF ablating branch model $\mathcal{F}_e$, only the original sentence \textit{X} and biased tokens \textit{B}, but not ensemble feature \textit{E}, directly affect the model decisions \textit{Y}. Therefore, TE of variables on \textit{Y} can be written as:

\begin{small}
\begin{equation}
TE = Y_{x,b}-Y_{x^*,b^*}.
\end{equation}
\end{small}And NDE of \textit{B} on \textit{Y} is:

\begin{small}
\begin{equation}
NDE = Y_{x^*, b}-Y_{x^*,b^*} .
\end{equation}
\end{small}We then obtain the TIE by comparing TE and NDE as debiased predicate logits:

\begin{small}
\begin{equation}
T I E=T E-N D E=Y_{x,b}-Y_{x^*, b}.
\end{equation}
\end{small}

Similarly, the causal graph of CCDF ablating $\mathcal{F}_x$ is shown in \figurename~\ref{fig:casual_ab}(b) and the causal effect is calculated as follows:

\begin{small}
\begin{equation}
TE = Y_{e,b}-Y_{e^*,b^*}.
\end{equation}
\end{small}

\begin{small}
\begin{equation}
NDE = Y_{e^*, b}-Y_{e^*,b^*} .
\end{equation}
\end{small}

\begin{small}
\begin{equation}
T I E=T E-N D E=Y_{e,b}-Y_{e^*, b}.
\end{equation}
\end{small}

\end{document}